\documentclass[runningheads]{llncs}
\usepackage{graphicx}
\usepackage{algorithm}
\usepackage[noend]{algpseudocode}

\begin{document}
\title{Making Metadata More FAIR Using Large Language Models}
\titlerunning{\scriptsize DaMaLOS@ESWC PUBLISSO-Fachrepositorium. DOI: 10.4126/FRL01-006444995}
%
 \author{Sowmya S. Sundaram\orcidID{0000-0002-0086-7582}
 \and
 Mark A. Musen\orcidID{0000-0003-3325-793X}
 }
%
\authorrunning{Sundaram and Musen (2023): Making Metadata More FAIR Using LLMs}
%
\institute{Stanford University}
%
\maketitle              
\begin{abstract}
With the global increase in experimental data artifacts, harnessing them in a unified fashion leads to a major stumbling block - bad metadata. To bridge this gap, this work presents a Natural Language Processing (NLP) informed application, called FAIRMetaText, that compares metadata. Specifically, FAIRMetaText analyzes the natural language descriptions of metadata and provides a mathematical similarity measure between two terms. This measure can then be utilized for analyzing varied metadata, by suggesting terms for compliance or grouping similar terms for identification of replaceable terms. The efficacy of the algorithm is presented qualitatively and quantitatively on publicly available research artifacts and demonstrates large gains across metadata related tasks through an in-depth study of a wide variety of Large Language Models (LLMs). This software can drastically reduce the human effort in sifting through various natural language metadata while employing several experimental datasets on the same topic.
\keywords{Metadata \and NLP \and FAIR data}
\end{abstract}
\section{Introduction}


Metadata are the description of data that scientists, in turn, use for analysis and other complex downstream tasks. However, if the metadata are not of good quality, it can hamper retrieval, querying, analysing and harnessing the publicly available digital research artifacts. This problem has been often cited in the literature (\cite{musen2022without,Dua2019}) and was the basis of the keynote talk in a previous set of proceedings of this workshop \cite{kramer2021open}. This formidable bottleneck cripples the use of digital artifacts (including experimental datasets, scientific data descriptions, ontologies etc.) in an interoperable manner. Metadata often have representational heterogeneity - that is different terms refer to the same value, which often do not conform to specifications. The datasets are simply not `AI-ready' if the metadata are not good. 

\begin{figure}
    \centering
    \includegraphics[width=\textwidth]{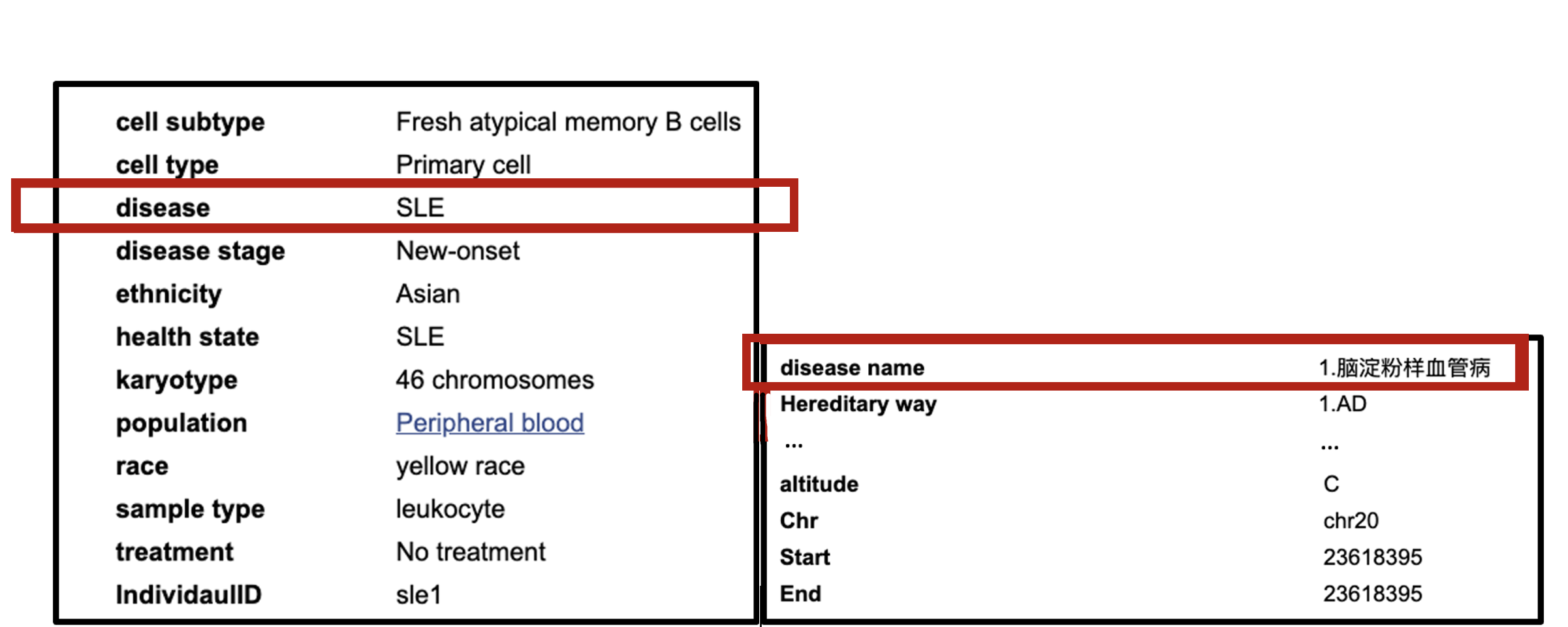}\\
    \includegraphics[width=\textwidth]{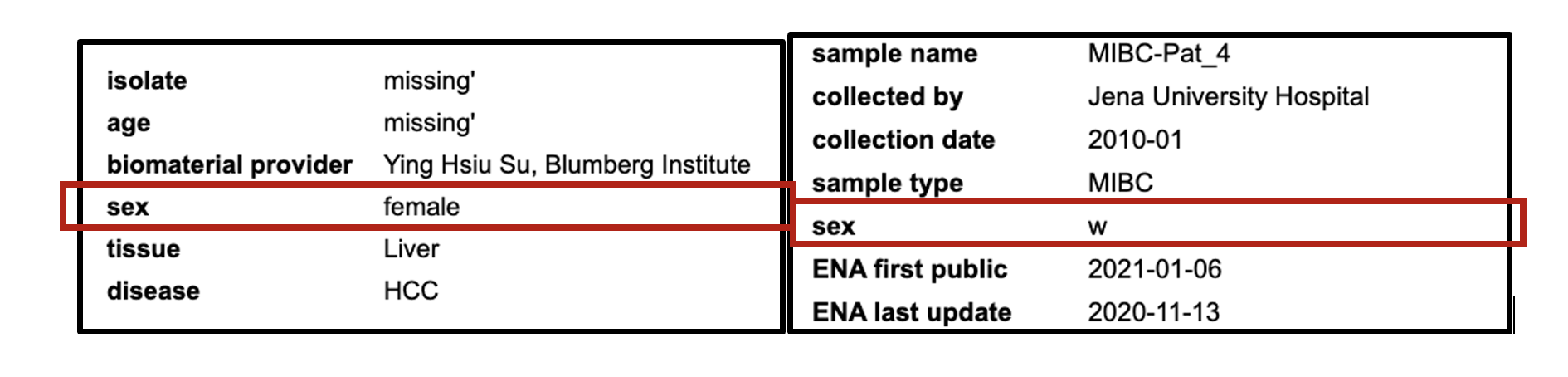}
    \caption{Real World Metadata - The red boxes bring attention to messy metadata in real records denoted by black boxes.}
    \label{fig:metadata}
\end{figure}

In a bid to enhance the sharing of scientific data, the FAIR (Findable, Accessible, Interoperable and Reproducible) principles \cite{wilkinson2016fair} were introduced. The purpose was to enable reuse and verify existing scientific data.  Though such principles are exemplary for \textit{future} scientific data artifacts, the mere presence of such guidelines does little to unify \textit{existing} artifacts across seemingly diverse but semantically similar resources

There is a long reported difficulty in finding datasets pertaining to a specific format or topic \cite{musen2022without}. Our group has been consistently working towards making research artifacts FAIR, especially in the arena of processing metadata. First, an in-depth empirical analysis was done that conclusively determined that the quality of metadata is poor in digital research artefacts \cite{gonccalves2019aligning}.  In this process, a robust software tool was developed - a FAIR Workbench (\cite{cedar}, \cite{Musen2022}) that can recommend cleaning up a research artefact before submitting it. 

FAIRMetaText was designed with principles deriving from an earlier research foray by our lab to incorporate NLP for automating metadata clean-up. The algorithm \cite{gonccalves2019aligning} used the NLP technology available then and performed metadata analysis in a semi-automated manner. The method handled spelling mistakes with difficulty and circumvented it by a syntactic similarity function. With the advent of Large Language Models (LLMs), NLP practitioners are able to make better strides and provide a completely automated approach, across many domains. Hence, in this work, we endeavor to upgrade our previous algorithm by (a) using improved NLP models and (b) reducing human intervention. In this work, an attempt to \textit{automatically} address some of the metadata issues of web artifacts has been made with the use of the state-of-the-art Natural Language Processing (NLP) tools for the \textit{existing} metadata. 

Experimental metadata are made of \textit{reporting guidelines} that specify the name of metadata and \textit{ontologies} that provide possible values. For example, while observing diseased samples, sex of the patient has been referred to variously as `F', `female' and `w'. Similarly, the disease is sometimes altogether missing, mentioned in abbreviations such as `SLE' or presented in multi-lingual terms. In the previous examples, the field names are `sex' and `disease' and the field values have been described. As seen in Figure ~\ref{fig:metadata}, `disease' is variously described as `disease' and `disease name'. FAIRMetaText can handle both field \textit{names} and \textit{values}.

Since the previous version of this algorithm differed significantly in its mechanism and was semi-automated, it has not been included in the empirical results. For the evaluation, both syntactic (spelling and phrase) and semantic similarities (meaning and context) of metadata are explored.

FAIRMetaText converts metadata into vectors based on the available large language models. These vectors can then be used in a wide variety of application. For our empirical study, there will be an exploration of two methods - retrieval and clustering. The experiments demonstrate that retrieval is useful for metadata compliance and that clustering is useful for metadata disambiguation through experiments. For this, both real and simulated datasets have been used. Through this tool, the authors propose that precious human effort applied to clean up metadata can be directed towards more useful endeavors.

\section{Related Work}

Metadata issues had been addressed time and again in the past (\cite{musen2022without,metadata_quality,Bruce2004metadatadefn,Bui_Park_2013_META}). The FAIR \cite{wilkinson2016fair} principles suggest that metadata must be Findable, Accessible, Interoperable and Reusable. This work addresses a way to use LLMs \cite{eloundou2023gpts} to make data FAIR.

There has been a lot of work in converting Knowledge Graphs, in their entirety, into vectorial embeddings for further processing \cite{wang2017knowledge}. This approach helps in question answering and other related tasks for structured data. This method is only useful where the metadata are specified perfectly. In this work, the focus is on developing embeddings for the NLP descriptions, rather than RDF triples or any other form of structural definitions. However, the presented algorithm is simply a text-based algorithm that can be applied across a wide variety of web artifacts.  

Metadata alignment where manual mappings such as SSSOM \cite{10.1093/database/baac035} are used to bridge the gap between web resources, is a popular approach in the community. However, our work is designed to push the limits of automatically cleaning up metadata using state of the art NLP techniques. 

Multiple research directions have been undertaken by our lab to enhance metadata authoring and recommendations (\cite{egyedi2017embracing,egyedi2018using,martinez2019using}). These methods harness advancements in machine learning and semantic technologies. To ensure the quality of metadata being submitted, \cite{Musen2022}, the FAIRware workbench highlights errors in submitted data with respect to the metadata specifications. The workbench ascertains and tests the quality of new scientific research being submitted.  FAIRMetaText is designed with the \textit{existing} metadata in mind by exploiting LLMs. There has been less work in this space of using NLP. As mentioned before, the earlier version \cite{gonccalves2019aligning} mapped possible candidates syntactically through edit distance and clustered semantic inputs using primitive word embeddings such as word2vec \cite{mikolov2013distributed} and GloVe \cite{pennington2014glove}. The method also involved a human in the loop. FAIRMetaText  combines all the automated steps using the rich embeddings provided by the LLMs vis-a-vis the static word embeddings explored earlier. Hence, our current method is more robust, powerful and completely automated.

\section{Large Language Models}

Text is converted into \textit{embeddings}, that is a piece of text is embedded in a vector space. Early naïve methods of embedding text into vector spaces included converting the word into a \textit{one-hot vector}. Then, static word embeddings were in vogue where a single word is converted into a vector using a lookup \cite{mikolov2013distributed}. This model fails when an unknown word is presented or when contexts change the meanings of words. The predecessor of this work \cite{gonccalves2019aligning}, relied on a mix of syntactic analysis followed by the use of such static embeddings. Now, LLMs use complex text embeddings that incorporate context and can handle unknown words. The evolution of embeddings along with their corresponding visualizations is presented in Table ~\ref{tab:embeddings}.

\begin{table}
\centering
\small
\caption{Evolution of Word Embeddings - The figures are word embeddings representation. The axes are of large vector dimensions projected on to two dimensions, hence the dimensions hold no meaning and are unlabelled}
\def\arraystretch{1.05}
\begin{tabular}{p{0.45\textwidth} p{0.55\textwidth}}
\hline
\hline
\multicolumn{2}{p{\textwidth}}{
Consider a dictionary of five words - `the', `a', `Greece', `metadata', `data' and the corresponding vectors $v_{the}, v_{a}, v_{Greece}, v_{metadata}, v_{data}$.}\\
\hline\parbox[c]{0.45\textwidth}{
\textbf{One-hot Embeddings:} Text is converted into unit vectors that are the length of the vocabulary. This leads to sparse vectors that are equidistant. All words are equidistant and hence does not capture semantics of similar words.}&
    \parbox[c]{0.55\textwidth}{\includegraphics[width=0.55\textwidth]{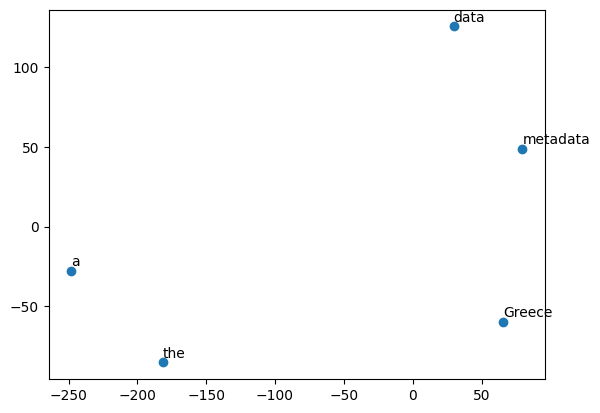}}
\\
\hline
\parbox[c]{0.45\textwidth}{\vspace{0.5em} \textbf{Frequency-based Embeddings:} Instead of one-hot embeddings, the weight of the word is offset by their frequency in their document and the general frequency across documents. This captures some aspects of importance of words in a document but still does not capture semantics completely. The vectors are still high-dimensional and sparse.} &\parbox[c]{0.55\textwidth}{\includegraphics[width=0.55\textwidth]{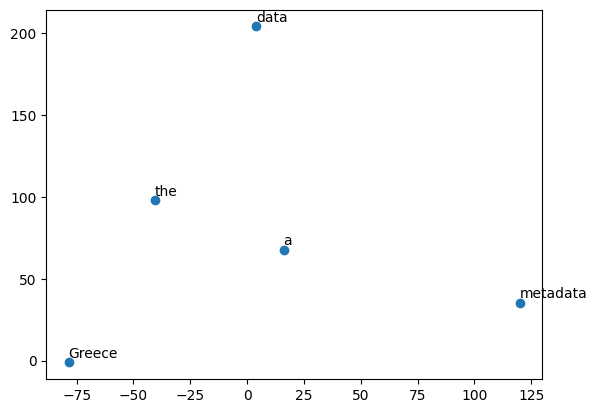}}\\
\hline
\parbox[c]{0.45\textwidth}{\vspace{0.75em}\textbf{Distributed Embeddings:} Using neural networks and large corpus of data, in this method of building distributed embeddings, the context of words are used to learn dense vectorial representations. This method produces relatively denser embeddings of length \~300. These embeddings can capture semantic relationships. In our example, the words `data' and `metadata' are closer. However, these embeddings cannot handle out-of-vocabulary words.\vspace{0.5em}} &\parbox[c]{0.55\textwidth}{\includegraphics[width=0.55\textwidth]{tfidf.png}}\\
\hline
\parbox[c]{0.45\textwidth}{\vspace{0.75em}\textbf{Transformer Embeddings:} Transformer based models obtain denser representations based on larger models and more data. These embeddings can handle out-of-vocabulary words as the vector is constructed from representations of sub-words rather than words.\vspace{0.5em}} &\parbox[c]{0.55\textwidth}{\includegraphics[width=0.55\textwidth]{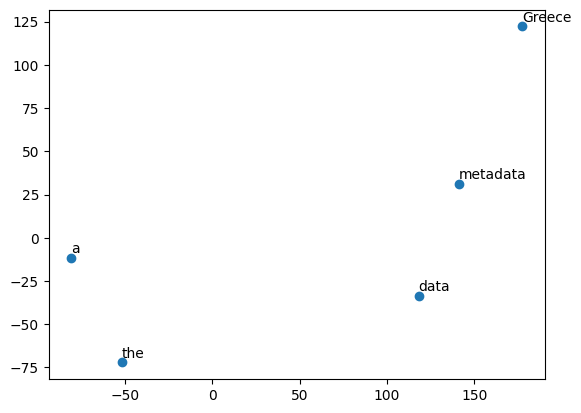}}\\
\hline
\end{tabular}
\label{tab:embeddings}
\end{table}

Large Language Models (LLMs) have taken up the world of NLP by storm \cite{eloundou2023gpts}. LLMs are trained over huge swathes of data (GPT3 uses a dataset of nearly 380 terabytes of data! \cite{gpt3}). They are instances of \textit{self-supervised learning} where the data in itself is enough to train the systems without costly annotation. Leveraging the freely available text online, large language models identify patterns in text and have been proven to be successful in generating text. Under the hood of this generation process is the presence of word embeddings, or rather text embeddings, that are learned as a by-product. Text embeddings convert a piece of text to a vector such that \textit{similar} pieces of text have vectors that are close to each other in distance. These models can be used on unseen data without further training for a particular application, which is known as zero-shot learning. Thus, it is a potent tool, in low resource data or data where annotation is hard, that has been harnessed for this application. Popular generic LLMs such as BERT, RoBERTa, CharBERT and PhraseBert have been used along with biologically specific LLMs such a BioMedLM and BioBERT have been tested. The last LLM used is GPT3.5. In this work, the embeddings \footnote{\url{https://platform.openai.com/docs/guides/embeddings/what-are-embeddings}} provided by OpenAI are plugged in directly, as they are more useful for our applications. 

\section{Applications of Embeddings to Metadata}
The following properties of embeddings have been exploited to process metadata (a) \textbf{Structural Similarity:} Text embeddings that have similar structure (such as spelling or character) are closer together in the vector space; and (b) \textbf{Semantic Similarity:} Text snippets that have similar meaning are closer together in vector space.

\subsection{Retrieval for Metadata Compliance}

Given a metadata term $w$ and a permissible list of ontological terms $\mathcal{D} = \{d_1, d_2, ... ,d_k\}$, if $w \notin \mathcal{D}$, then $w$ is said to be non-compliant. For compliance, the task is to retrieve $d \in \mathcal{D}$ such that the cosine similarity between the vector forms generated by model $m$ to obtain $vec(d)$ and $vec(w)$, is maximum $\forall d in \mathcal{D}$. For evaluation, a check against a specified ground truth of which ontological term $d_{true} \in \mathcal{D}$ matches is performed. Then, accuracy is calculated based on the number of matches across $n$ data points $\sum{d = d_{true}}/n$.

\subsection{Clustering for Metadata Unification}

Given a set of metadata terms $\{t_1, t_2,...,t_n\} \in \mathcal{T}$, it is converted into a set of vectors $\{v_1, v_2,...,v_n\} \in \mathcal{V}$. Then, the popular clustering algorithm k-means \cite{kmeans} is employed to group them into clusters $\mathcal{C}$. For evaluation, the purity metric is used (by labeling each cluster with a majority vote) to measure against a set of cluster labels that were not used as part of the algorithm. Assume that the true cluster assignments $\mathcal{C_{true}}$ are known beforehand for these terms. For each cluster $c \in \mathcal{C}$, determine the majority label of the data points according to $\mathcal{C_{true}}$ Then, measure the fraction of data-points that adhere to this labelling. the experiments also involve qualitative examination of the clusters by projecting the vectors onto a two dimensional space using visualisation algorithms like t-SNE \cite{van2008visualizing}.

\begin{algorithm}[h]
\footnotesize
\caption{Metadata Analysis}\label{alg:met}
\begin{algorithmic}[1]
\Procedure{GetVectors}{$d$, $D$, $m$}
\State $\mathcal{V} \gets \emptyset$
\For{$d$ $\in$ $\mathcal{D}$}\Comment{For every word $d$ in dataset $\mathcal{D}$}
\State $v_d \gets m(d)$ \Comment{Get vector generated by $m$ for $d$}
\State $\mathcal{V} \gets \mathcal{V} \cup v_d$
\EndFor
\State \textbf{return} $\mathcal{V}$
\EndProcedure
\Procedure{MetadataCompliance}{word $w$, accepted terms $\mathcal{D}$, model $m$}
\State $sim \gets$ 0
\State $\mathcal{V} \gets $ \Call{GetVectors}{$d$, $D$, $m$}
\For{$v$ in $\mathcal{V}$}
\State $sim_w$ $\gets$ \Call{CosineSimilarity}{$w$, $v$}
\State $sim_d$ $\gets$ $\emptyset$
\If $sim_w > sim$
\State $sim \gets sim_w$
\State $sim_d \gets d$ \Comment{Storing the word $d$ corresponding to $v$}
\EndIf
\EndFor
\State \textbf{return} $sim_d$
\EndProcedure
\Procedure{MetadataUnification}{metadata terms $\mathcal{T}$, model $m$, $k$}
\State $\mathcal{V} \gets $ \Call{GetVectors}{$d$, $D$, $m$}
\State $clusters$ $\gets$ \Call{KMeans}{$V$, $k$}
\State \textbf{return} $clusters_d$ \Comment{Clusters of words corresponding to $clusters$}
\EndProcedure
\end{algorithmic}
\end{algorithm}

\section{Methods}

In this section, the methods of evaluating two tasks are presented - (i) Metadata compliance and (ii) Metadata Unification. The datasets and evaluation metrics are described along with the performance of FAIRMetaText on these tasks.

\subsection{Retrieval for Metadata Compliance}
For this task, three datasets have been chosen. To demonstrate the generalisability of our approach, two web artifacts - the Adult and Mushroom datasets are explored. These datasets were chosen from the UC Irvine Machine Learning Repository \cite{Dua2019} as they had a large number of text categorical variables and can be easily extrapolated to \textit{any} digital experimental data. The third dataset is the authors' own in-house expert-curated dataset on tissue samples. The dataset values are derived from HuBMAP specifications \cite{hubmap} and expert introduced errors. These datasets have a large number of categorical values. Also, these experiments involve only the dataset description in text and not the actual samples.

With the Adult and Mushroom datasets, the dataset specifications are perturbed by single character substitutions using a publicly available NLP augmenter \cite{ma2019nlpaug}. The NLP augmenter is a piece of software that introduces errors into words based on  user requirement. Hence, this process generated the two new datasets after adding single character errors. These simulated datasets are used for evaluation. A single character perturbation can be easily rectified using Levenshtein distance \cite{levenshtein}. However, with no training, this experiment demonstrates the efficacy of this method on many other facets through varied experiments. This experiment, as demonstrated in Table ~\ref{tab:syntactic}, measures the ability of the algorithm to identify syntactic similarities. 

\begin{figure}
    \centering
    
    \includegraphics[width=0.7\linewidth]{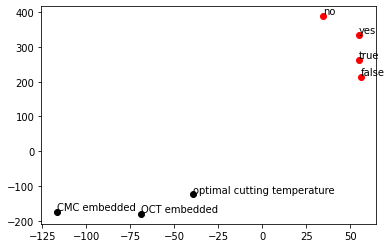}
    \caption{\small Visualising Semantic Similarity of Embeddings: The vectorial representations of seven terms have been projected onto two dimensions and presented here. Given two types of terms - domain specific terms and a set of Boolean terms, the clustering algorithm correctly identifies the semantic categories. Furthermore, the full form `optimal cutting temperature' is close to the phrase `OCT embedded'.}
    \label{fig:tissue}
\end{figure}

With the Tissue Sample dataset, users give a lot of semantic equivalences which are syntactically different as the wrong input. This phenomenon has been visualised in Figure ~\ref{fig:tissue}. The input was two sets of both user input and metadata specification. The vectors are of size 784 and to visualise them, they were projected onto two axes using t-SNE \cite{van2008visualizing}. The clustering algorithm correctly identified the diversity in text. In the figure, in the left bottom corner, the metadata specification `OCT embedded' is closest to the user term optimal cutting temperature. These semantic details are hard to capture in traditional metadata tools. 

The accuracy results for two settings - with and without definition for the expert curated dataset is presented in Table ~\ref{tab:real_data}. Without definition, the specification terms $\mathcal{D}$ are simply converted into vectors and compared against the query term $w$. In the other setting, the specification terms are augmented by concatenating their definitions as well. Hence, $\mathcal{D}$ will become $\{d_1+defn(d_1), d_2+defn(d_2), ... , d_k+defn(d_k)\}$. The NLP models will output a vector for these new chunks of text of size 784, that can then be used for the accuracy analysis reported in Table ~\ref{tab:real_data}. The number of samples was small (62), again a highlight of the difficulty of low-resource datasets.

\begin{table}
\begin{minipage}[c]{0.5\textwidth}
    \centering
    \small
    \caption{Metadata Compliance Suggestion on Simulated Data: Accuracy of retrieval in percentage, dataset size in Parenthesis}
    \renewcommand{\arraystretch}{1.25}
    \begin{tabular}{c |c |c }
        \hline
        \textbf{Model} & \textbf{Adult} & \textbf{Mushroom} \\
        & \textbf{[808]} & \textbf{[2222]} \\
        \hline
        \hline
         BERT & 46.71 & 45.19 \\
         ROBERTA & 56.14 & 68.09 \\
         CharBERT & 43.00 & 47.14\\
         PhraseBERT & 47.71 & 51.09 \\
         GPT & \textbf{87.78} & \textbf{86.95}\\
        \hline
    \end{tabular}
    \label{tab:syntactic}
\end{minipage}\hfill
\begin{minipage}[c]{0.45\textwidth}
\centering
   
    \caption{Metadata Compliance on Real Data: Accuracy of retrieval in percentage - with and without the use of  definition of metadata}
    \small
    \begin{tabular}{c |c | c}
       \hline
        \textbf{Model} & \textbf{No Defn} & \textbf{With Defn} \\
        \hline
        \hline
         BERT & 58.70 & 45.65\\
         ROBERTA & 52.17 & 45.65 \\
         CharBERT & 52.17 & 41.30 \\
         PhraseBERT & 54.35 & 54.34\\
         BioBERT & 58.69 & 56.52\\
         BioMedLM & 47.82 & 45.65\\
         GPT & \textbf{63.04} & \textbf{67.39} \\
         \hline
    \end{tabular}
    \label{tab:real_data}

\end{minipage}
    
\end{table}

Among all the LLMs, the GPT embeddings outperform with a large margin. The improved results with definitions suggests that with good representation of meanings of categorical specifications, text algorithms can do better. This observation is a pertinent point for the collection of metadata in future.

\subsection{Clustering for MetaData Unification}

For these experiments, the sources include the BioSample \cite{biosample} resource which is a database maintained by the U.S. National Center for Biotechnology Information (NCBI) that provides descriptions and metadata for biological samples used in research. In BioSample, there is a list of attributes with their synonyms that can be used for simulating the clustering task. As already mentioned, clustering the vectors will lead us to discover groups of similarly meaning metadata terms. Purity will be used to measure the clustering. Since the knowledge of synonyms is known beforehand, one can evaluate quality of the clustering algorithm that had been presented with unlabelled metadata terms. In Table ~\ref{tab:clustering}, the experiments have been presented by the outcome of the k-means clustering algorithm on BioSample synonym dataset at k = 100, 200 and 500 on a set of 1500 terms.

\begin{table}[]
    \centering
    \small
    \caption{Metadata Unification: Purity of clustering metadata terms and their synonyms using k means for varying k }
    \renewcommand{\arraystretch}{0.75}
    \begin{tabular}{l |c  c  c }
       \hline
        \textbf{Model} & \textbf{Purity (k=100)}& \textbf{Purity (k=200)} & \textbf{Purity (k=500)}\\
        \hline
        \hline
         BERT &  59.36 & 59.40 & 57.18 \\
         ROBERTA & 58.83 & 58.70 & 57.90 \\
         CharBERT & 61.01 & 61.54 & 60.82\\
         PhraseBERT & 64.99 & 64.46 & 61.48\\
         BioBERT & 58.90 & 59.30 & 58.30\\
         BioMedLM & 59.03 & 59.03 & 58.10\\
         GPT & \textbf{79.48} & \textbf{75.51} & \textbf{72.99}\\
         \hline
    \end{tabular}
    \label{tab:clustering}
\end{table}

\begin{figure}[h!]
    \centering
    \includegraphics[width=0.7\textwidth]{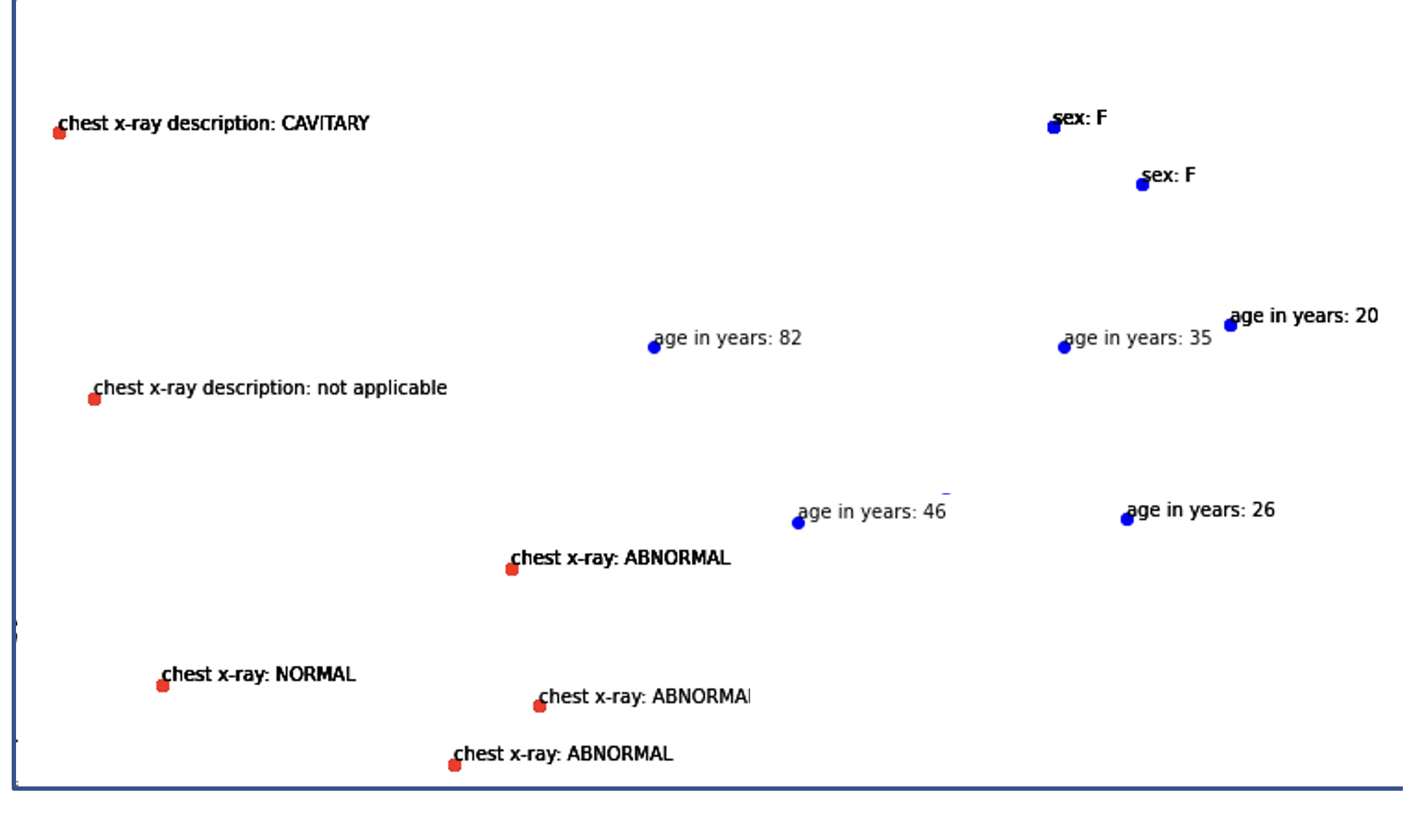}
    \caption{\small Clustering of Personal vs Diagnostic Entities - Two dimensional projection of embedding vectors of metadata fields and values. Red terms correctly clustered diagnostic metadata and purple terms clustered patient metadata. The axes are a projection of large vector dimensions on to a 2D space and hence are unlabelled.}
    \label{fig:tb2}

\end{figure}

Qualitatively, the NCBI's GEO (U.S. National Center for Biotechnology Information’s Gene Expression Omnibus) database has been used to examine clusters more closely. GEO is a public database that provides access to a large collection of gene expression data. The GEO database includes data from a variety of platforms, such as micro-arrays, next-generation sequencing, and gene expression profiling assays. The text descriptions in the 'characteristics' column of all tuberculosis samples were extracted and clustered. This process was done using the R package 
GEOMetaDB \cite{geometadb}. With a similar approach of projection onto two dimensions using t-SNE, the findings are presented in Figure ~\ref{fig:tb2}. As can be judged from the images, similar terms lie close to each other in the vector space. Given below in Figure ~\ref{fig:age} is another experiment where all `age' related terms were clustered. These experiments describe how text information within metadata can be harnessed for practical applications of compliance and unification.

\begin{figure}[h]
    \centering
    \includegraphics[width=0.8\textwidth]{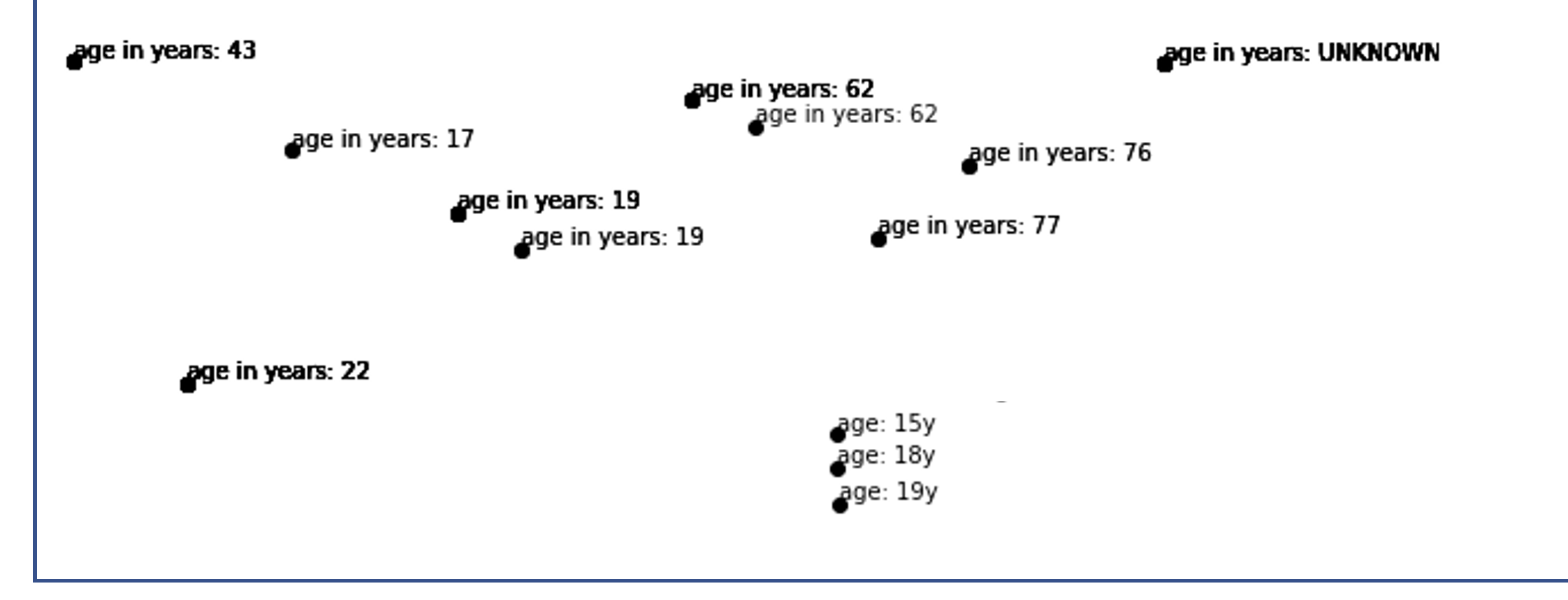}
    \caption{\small Clustering of `Age' Related Terms - Two dimensional projection of embedding vectors of metadata fields and values relating to `age'. Different configurations of the term `age' are clustered into the same bin - whether they are `age = [X]y' or `age in years: [X]'. The axes are a projection of large vector dimensions on to a 2D space and hence are unlabelled.}
    \label{fig:age}
\end{figure}


\section{Discussion}
Metadata are the often underappreciated aspects of experimental datasets. This approach is a roadblock to efficient usage of scientific artifacts. While one can postulate theories to improve the collection of such web artifacts in future, there is a pressing need to grapple with the existing huge mass of poor quality of metadata \cite{metadata_quality}. 

This work described FAIRMetaText which can help in this requirement. FAIRMetaText is an NLP based tool that can analyze the textual descriptions of metadata and help immensely in cleaning up the low quality data. Scientists spend a lot of time trying to look up relevant pieces of information on datasets that are simply hard to access. FAIRMetaText can cut their searching time drastically. Since LLMs encode complex semantic information, they can be leveraged to give text embeddings that can model metadata with no user/domain specific training. These embeddings were shown to capture both syntactic and semantic similarity of metadata. This study was done with an empirical analysis with different LLMs. Surprisingly, the generic GPT model was found to perform better than models trained especially for bio-medicine in the case of medical metadata. One possible surmise for this is that this is because of the much larger data these generic models have been trained on, that can account for vagaries in form, spelling and semantics. 

\begin{figure}
    \centering
    \includegraphics[scale=0.3]{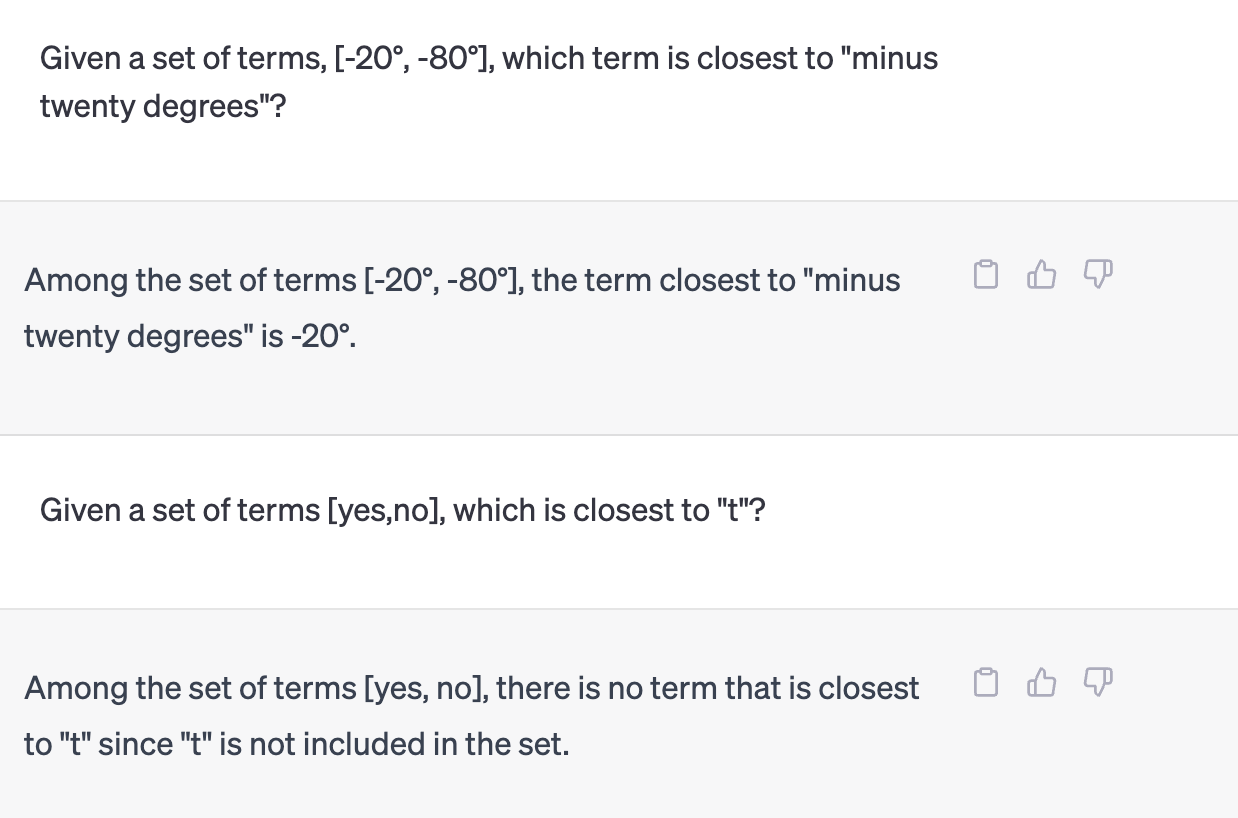}
    \caption{ChatGPT and Metadata Compliance}
    \label{fig:chatgpt}
\end{figure}

A natural question at this juncture is whether doing away with FAIRMetaText and directly employing ChatGPT can help. ChatGPT has a strict text-to-text interface (Fig ~\ref{fig:chatgpt}). FAIRMetaText goes under the hood and employs embeddings in a way suitable for metadata analysis. For example, clustering over 20000 terms would be tedious. Hence, FAIRMetaText builds on top of GPT embeddings and is tailored for metadata clean-up.
Using FAIRMetaText, (a) one can monitor the quality of data and metadata of research artefacts being submitted online and (b) analyze and compare existing metadata to make artefacts inter-operable. This tool can be used on \textit{any} metadata term - both field names and field values. The prototype will be released soon. The requirement for FAIRMetaText to enforce metadata compliance is a set of \textit{accessible} metadata specifications that can be used to model the metadata terms. For metadata unification, the text descriptions of metadata must be machine accessible - easily downloadable and easy to parse. In future iterations of this product, FAIRMetaText will involve parsing various formats such as `json' and `xml' and `text' files for generalised processing.      

When this model was tested on the real data requirements, the GPT model, though best, could only provide a retrieval accuracy of ~60 percent. This suggests that further training or \textit{finetuning} of these LLMs for the purpose of metadata analysis could potentially improve the performance. User data and information of metadata non-compliance can also be logged in future to enhance the model. The experiments show that, despite the shortcomings, there is an enormous potential to be harnessed as the formidable results from generic models suggest. Unlike traditional deep learning models that require large amounts of data, just a fraction of that size is enough for \textit{fine-tuning} or enhancing the embeddings with domain-specific information. This investigation leads to a conclusion that indeed, FAIRMetaText is a significant and promising direction for \textit{automatically} making varied \textit{existing} messy metadata FAIR and for specifying FAIR compliant metadata suggestions for future experimental datasets.

\section{Conclusion}
To address concerns of the quality of \textit{existing} metadata, the authors presented an automatic metadata analysis and improvement tool. In accordance with the FAIR principles, the power of Large Language Models was used for \textit{automatically} analyzing metadata. The current text embedding methods fare better than the erstwhile word embeddings \cite{gonccalves2019aligning} due to the following properties, namely identifying both syntactic and semantic properties congruently and the ability to handle out of vocabulary texts. The efficacy of our algorithms were demonstrated for metadata compliance and unification. These experiments were performed on both simulated and real datasets . In future work, the authors look to analyse actual user logs for metadata compliance and develop annotated datasets for evaluating the algorithms better. Fine-tuning LLMs with domain-specific data has been shown to improve performance in the literature \cite{eloundou2023gpts}. In this work, commercial GPT embeddings have been used. An exploration of open-source versions of the same forms the basis of the next increment of this tool. 

\section{Acknowledgments}
This work was supported in part by grant R01 LM013498 from the U.S. National Library of Medicine and by Award OT2 OD033759 from the U.S. National Institutes of Health Common Fund. The authors thank Josef Hardi for providing the expert-curated dataset and extensively working on the metadata of HuBMAP \cite{hubmap}. This project would not be possible without the accesible data available on BioPortal \cite{bioportal} and NCBI's GEO \cite{geometadb}.

\bibliographystyle{splncs04}
\bibliography{mybibliography}
%




\end{document}